\documentclass[letterpaper, 10 pt, conference]{ieeeconf}  

\IEEEoverridecommandlockouts                              

\usepackage{graphicx}

\usepackage{graphics} 
\usepackage{epsfig} 
\usepackage{mathptmx} 
\usepackage{times} 
\usepackage{amsmath} 
\usepackage{amssymb}  
\usepackage{float}
\usepackage{hyperref}
\usepackage{amssymb}
\usepackage{pifont}
\usepackage[table]{xcolor}
\newcommand{\cmark}{\ding{51}}%
\newcommand{\xmark}{\ding{53}}%
\newcolumntype{C}[1]{>{\centering\arraybackslash}m{#1}}

\usepackage{color}
\usepackage{booktabs}

\usepackage{multirow}
\usepackage{comment}

\title{\LARGE \bf
Human-Robot Interaction using VAHR: \\ Virtual Assistant, Human, and Robots in the Loop
}

\author{Ahmad Amine$^{1}$, Mostafa Aldilati$^{1}$, Hadi Hasan$^{1}$, Noel Maalouf$^{2}$, and Imad H. Elhajj$^{1}$
\thanks{$^{1}$Authors are with the American University of Beirut, Lebanon \newline \tt \{aaa188, mad51, hsh24\}@mail.aub.edu, ie05@aub.edu.lb}
\thanks{$^{2}$Author is with the Lebanese American University, Byblos, Lebanon \tt noel.maalouf@lau.edu.lb}}

\begin{document}

\maketitle
\thispagestyle{plain}
\pagestyle{plain}

\begin{abstract}

Robots have become ubiquitous tools in various industries and households, highlighting the importance of human-robot interaction (\textbf{HRI}). This has increased the need for easy and accessible communication between humans and robots. Recent research has focused on the intersection of virtual assistant technology, such as Amazon's Alexa, with robots and its effect on HRI. This paper presents the Virtual Assistant, Human, and Robots in the loop (VAHR) system, which utilizes bidirectional communication to control multiple robots through Alexa. VAHR's performance was evaluated through a human-subjects experiment, comparing objective and subjective metrics of traditional keyboard and mouse interfaces to VAHR. The results showed that VAHR required 41\% less Robot Attention Demand and ensured 91\% more Fan-out time compared to the standard method. Additionally, VAHR led to a 62.5\% improvement in multi-tasking, highlighting the potential for efficient human-robot interaction in physically- and mentally-demanding scenarios. However, subjective metrics revealed a need for human operators to build confidence and trust with this new method of operation.
\end{abstract}

\section{INTRODUCTION}
Virtual assistants have become ubiquitous in our daily lives, integrated into smartphones, in-car infotainment systems, and smart speakers such as Google Home and Amazon Alexa. At the same time, robots are increasingly used for various tasks, from vacuuming homes \cite{roomba} to delivering goods in hospitals \cite{aethon}. The potential benefits of integrating virtual assistants and robots are thus increasingly appealing, as virtual assistants can enhance robots' physical capabilities, while robots can provide a natural Human-Robot Interface (HRI) for virtual assistants. In this paper, we propose VAHR (Virtual Assistant, Human, and Robots in the loop), a system that connects virtual assistants to robots through different communication schemes, either directly or indirectly. Our implementation uses Amazon's Echo Dot and Alexa virtual assistant, given their widespread availability and development environment (Alexa Skills Kit and AWS), but our proposed framework can be easily adapted to other virtual assistants. We review related work in Section~\ref{sec:rel_work} and describe our system design in Section~\ref{sec:design}. We then detail the implementation and experimental setup in Section~\ref{sec:implementation} before presenting the results and analysis in Section~\ref{sec:results}. The conclusion and future work are covered in Section~\ref{sec:conclusion}.

\section{Related Work} \label{sec:rel_work}
Recent research has focused on developing natural speech interfaces for human-robot communication using affordable home automation tools like Amazon Alexa. In this section, we will present various approaches to human-robot voice interfaces found in the literature, followed by an explanation of the commonly used HRI standard metrics for evaluating our proposed experiment.

\subsection{Voice-controlled Robots}
Earlier attempts of controlling robots through voice date back to 1998 and 2000 in which robots assisted mainly in surgery while responding to a limited set of voice commands and requiring the surgeon to wear a headset \cite{surgery_voice, 1998_surgery_voice}. 

Recently, robotic assistants such as Lio were deployed in hospitals to assist in healthcare, and disinfection \cite{lio}. Because of Lio's computational power, its voice interactions were purely synthesized natively without requiring any cloud-based solution \cite{lio}. However, these native solutions are considerably more expensive than cloud-based ones. Hence, controlling robots via a friendly communication scheme has been shifting towards low-cost cloud solutions. This particular field has attracted many applications in both industry and academia. In \cite{echobot}, researchers developed EchoBot, an interface between YuMi, the industrial robot, and Amazon Alexa to facilitate data collection for Learning from Demonstration (LfD) tasks. To test the efficacy of the experiment, a user is tasked with guiding YuMi to solve the Tower of Hanoi puzzle while occupying both hands. The user utters voice commands to the Echo Dot device to record the state of the two grippers involved in carrying out the task (open or closed) \cite{echobot}. This interface
was juxtaposed with a keyboard interface, and the results demonstrated that EchoBot was significantly more efficient than the keyboard interface \cite{echobot}. Alexa was also used as a speech interface in \cite{jungbluth_siedentopp_krieger_gerke_plapper} to control the Kuka robotic
arm with a gripper. The arm would respond to voice commands issued by the user to Amazon Alexa, and Alexa relays the command to the robot 
to pick and place a wrench from a toolbox via subscription to an MQTT topic \cite{jungbluth_siedentopp_krieger_gerke_plapper}. 


Further, in \cite{hri_iot}, a 3D-printed humanoid torso was manipulated through Alexa to enhance trust in
HRI. Additionally, as detailed in \cite{arm}, a six-degrees-of-freedom robotic arm was voice controlled through Alexa via a locally-hosted web server on Ngrok. Ngrok allows a user to host a local server with minimal effort. In \cite{auto-vehicle}, the motors of a
semi-autonomous lawn mower were strictly controlled through the Amazon Tap speaker, a voice automation tool equipped with Alexa. 

A more abstract work, such as in \cite{voice_alexa_platform}, depicts multiple approaches and architectures to interface one's robot with Alexa. To our knowledge, no work in the literature includes an architecture that allows for multi-robot bidirectional communication and is assessed according to standard HRI metrics. Our proposed VAHR system is assessed according to the metrics in the following subsection.   

\subsection{Popular HRI Metrics}
A comprehensive literature survey of existing HRI metrics over the past two decades was conducted by Murphy et al.~\cite{comprehensive}. It can be argued that the most standardized and cited metrics are summarized in \cite{hri-famous}. Saleh et al.~\cite{hri-metrics-general} improve these metrics. We list the metrics that are relevant to our study in what follows.

\begin{itemize}
     \item \textbf{Robot Attention Demand (RAD):} this metric measures the fraction of time spent by the user interacting with the robot \cite{hri-metrics-general}. Typically, This metric is calculated by the following formula:
     \begin{equation}
    \label{eq:RAD}
      RAD = \frac{IE}{IE + NT}   
     \end{equation}
      
     where NT represents the neglect tolerance, a metric that captures the time elapsed after the human operator issues the robot command until there is a notable drop in robot performance. IE denotes the interaction effort corresponding to the required time to interact with the robot \cite{hri-metrics-general}. Both of the metrics are acquired experimentally.
     \item \textbf{Fan-out (FO)} evaluates the efficiency and simultaneity of controlling multiple robots. This usually depicts the operator's efficiency in operating multiple homogeneous robots. It is computed by dividing the total task time by the RAD for a single robot.
     \begin{equation}
        FO = \frac{Total Task Time}{RAD}    
     \end{equation}

     \item \textbf{Trust:} to quantify the trust established between a human and a robot, a fuzzy temporal model is incorporated in \cite{hri-metrics-general}. This model takes quantifiable inputs such as fault size, productivity, and awareness and provides five states of trust as output (Very Low, Low, Medium, High, and Very High). Each of these states is associated with a value.
        (Very Low $\longrightarrow$ 0.1,
        Low $\longrightarrow$ 0.3,
         Medium $\longrightarrow$ 0.5,
         High $\longrightarrow$ 0.7, and
         Very High $\longrightarrow$ 0.9).
     The claim in \cite{hri-metrics-general} is that trust can significantly affect RAD. For instance, a human who is less trustworthy of his or her robot is likely to give greater attention to it even after sending the command for it to perform independent work. Hence, the following formula comes about according to \cite{hri-metrics-general}
     \begin{equation}
     RAD = DIT + IIT = DIT + NT*(1-T_r)    
     \end{equation}
     
     where DIT is the direct interaction time which is essentially the formula of (\ref{eq:RAD}), indirect interaction time (IIT) is a result of trust (\texorpdfstring{T\textsubscript{r}})) and neglect tolerance (NT). In this way, (3) captures the degree of trust the operator gives to the commanded robot. In controlled experiments, the value of trust can also be obtained through questionnaires rather than relying on a complex fuzzy model. 
     
     \item \textbf{Degree of mental computation:} some tasks may require the operator to significantly stress his/her short and long-term memory. Steinfeld et al.~\cite{hri-famous} mention examples of mental labeling of objects, object-referent association in working memory, and teleportation tasks. One may resolve this by resorting to feedback from the robot itself. One of the aims of this project is to allow robots to report feedback through Amazon Alexa. This metric is also useful in scenarios to assess controlled experiments. A popular tool named NASA-TLX (task load index) is typically utilized to measure this metric in controlled human experiments. It targets six fields in mental and physical workload  \cite{NASA}.
    
    
    \item \textbf{Efficiency and Effectiveness:} this is a common metric used in HRI \cite{hri-famous}. Efficiency measures the time needed to complete a task, while effectiveness is the success rate. It calculates the percentage of the robot's mission that was successful. 
 \end{itemize}

\section{System Design} \label{sec:design}

\subsection{Alexa to Robot Interaction}
Having surveyed the different approaches used in the literature, we will discuss these architectures in light of our approach to the problem.

\subsubsection{\textbf{MQTT Commands with Shadow Feedback}}
\begin{figure}[H]
    \centering
    \includegraphics[width=.48\textwidth]{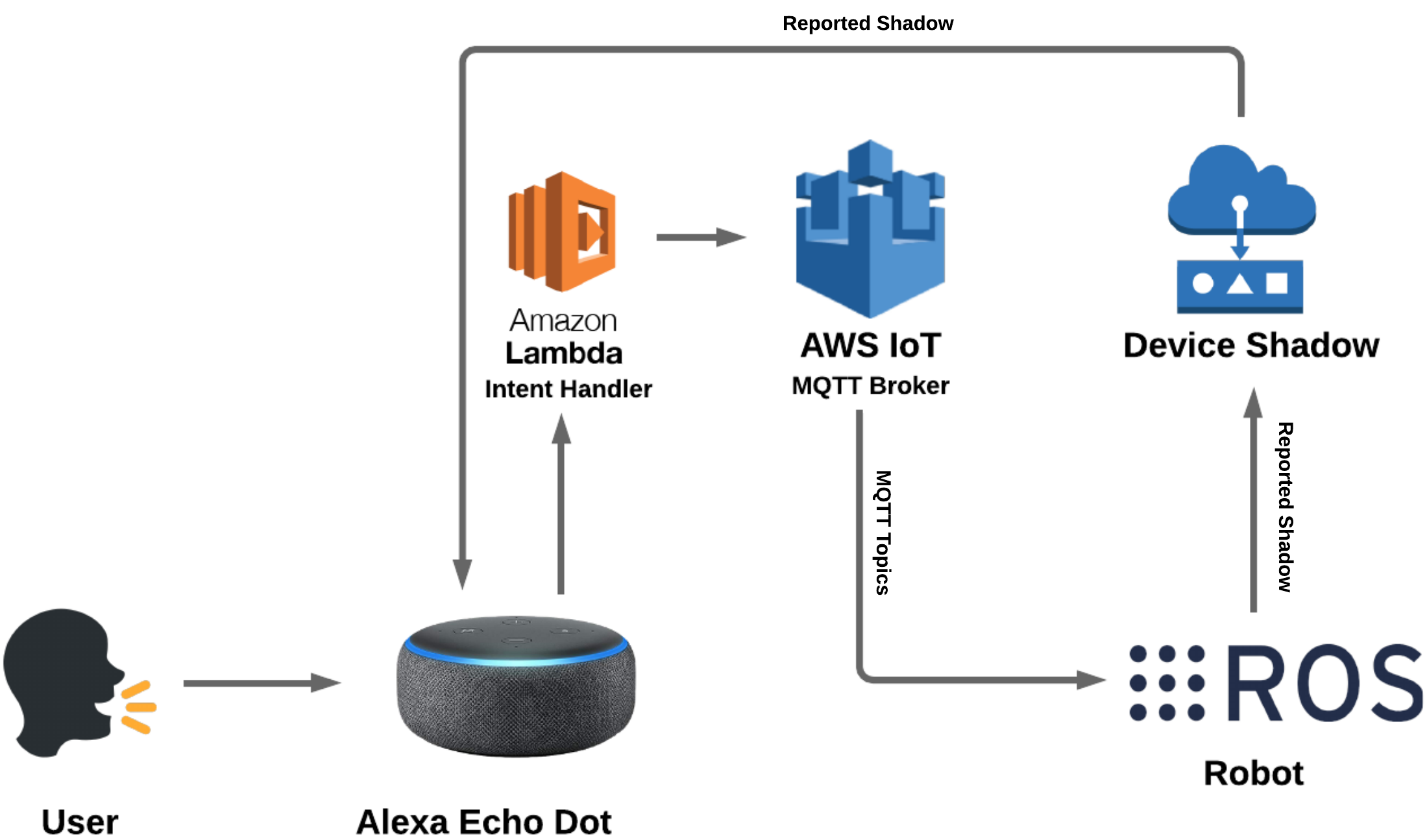}
    \caption{Our proposed communication scheme utilizes MQTT commands and the device shadow service.}
    \label{virtual robot 8}
\end{figure}
In this communication scheme, we combine the work done in \cite{hidalgo2019integration} with the work done in \cite{jungbluth2018combining} while abstracting the MQTT broker functionality to AWS. We thus utilize AWS IoT Core's MQTT broker to issue commands to robots while providing feedback to users through Alexa via the device shadow service, as shown in Fig.~\ref{virtual robot 8}. Relying only on device shadows as in \cite{jungbluth2018combining} incurs redundant shadow requests that increase the cost of the system. Instead, we minimize the number of shadow requests by using MQTT messages to enable asynchronous messages with one-to-many communication capability. This ensures efficient scaling of the communication architecture as more robots are added to the system.
\subsubsection{\textbf{One-to-Many Communication}}
     With MQTT messages, we can simultaneously broadcast a message from Alexa to several robots and devices using the publish-subscribe model. By subscribing to a common MQTT topic, multiple robots can receive commands in parallel as Alexa publishes these commands. This offloads communication from Lambda to the MQTT broker and thus ensures timely responses from Lambda. This also allows us to scale to swarms of robots if needed.
\subsubsection{\textbf{Asynchronous Communication}}
    Instead of requiring robots to actively query their shadows for commands, we initiate an MQTT subscriber, which listens to MQTT messages received at a certain port of the robot. This frees the robot from wasting computations and bandwidth to poll its device shadow continuously, especially when most requests are redundant. Now, the robot would only process requests when a message is received.
    
We can now describe the communication flow as follows.
\begin{enumerate}
    \item User invokes desired robot-related Alexa skill.
    \item Desired Alexa Skill interprets the spoken utterance and sends the interpreted intent to the intent handler endpoint.
    \item The AWS Lambda function receives the intent as an Alexa Skill intent handler.
    \item If the invoked intent requires a robot command to be initiated, AWS Lambda publishes a message with the command values to the respective MQTT topic (i.e., spin topic to spin a robot).
    \item All robots subscribed to that topic receive the message with its value when Lambda publishes to that topic. 
    \item After going through the intent logic and issuing the requested MQTT message (if needed), the intent handler requests the robot's shadow to receive the reported states by the robot (i.e., the command received successfully).
    \item The handler finally responds with a speech string to be uttered to the user by the Alexa Skill model, including any updates of the robot's states, if they exist.
    \item On the robot side, the device shadow is updated whenever any of the robot's states have changed (i.e., Task Completed or Stuck).
\end{enumerate}

Accordingly, we have created a scalable architecture that allows robots to integrate robustly and efficiently with Alexa. The advantages and disadvantages of this service can be summarized as follows.

\begin{itemize}
\item\textbf{Advantages}
    \begin{enumerate}
        \item \textbf{Efficient Communication:} Fewer shadow requests and MQTT messages are required.
        \item \textbf{Flexibility:} Enables one-to-many and one-to-one communication between Alexa and Robots.
        \item \textbf{Robustness:} The asynchronous nature of MQTT messaging, coupled with the robustness of the AWS MQTT broker, ensures that the system continues to function if any of the involved endpoints stops working. Commands and states would be delayed, but the communication is not fatally broken.
    \end{enumerate}
\item \textbf{Disadvantages} 
    \begin{enumerate}
        \item \textbf{Complexity:} Two services must be configured (MQTT Broker and device shadows) for this setup.
    \end{enumerate}
\end{itemize}
\subsection{Robot to Alexa Voice Interaction}

To further validate the importance of using Alexa in robotics, a system is devised that grants a robot the ability to communicate with Alexa through speech. This system is constrained to the purposes of the experiment explained in the next section, but it can be further extended to be deployed in any application. In essence, the robot utters commands that trigger Alexa to respond, and according to this response, the robot performs a designated action sequence. 

\begin{figure}[H]
    \centering
    \includegraphics[width = 0.45 \textwidth]{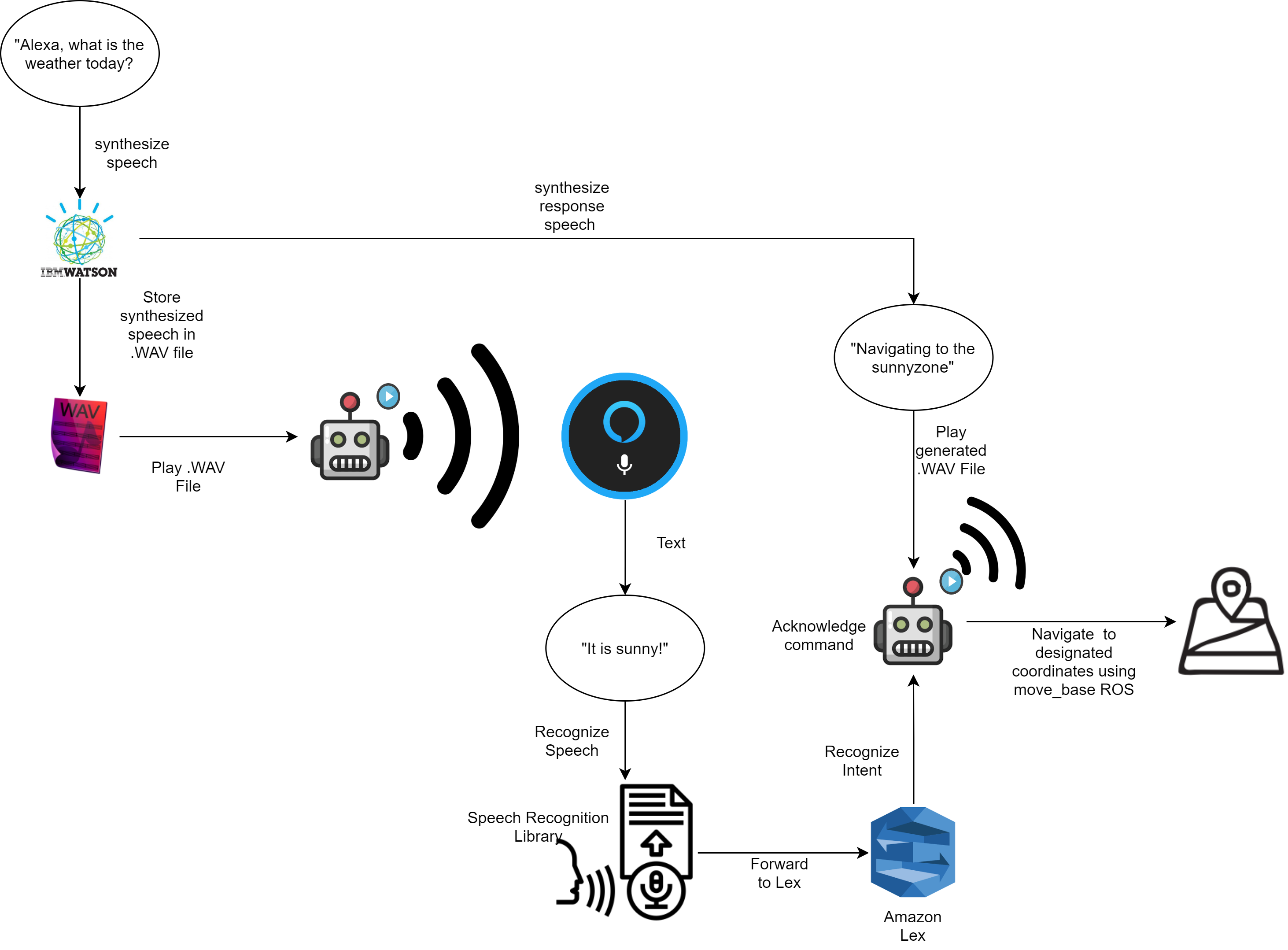}
    \caption{Robot to Alexa voice System architecture as applied to a weather application.}
    \label{weatherbot}
    \vspace{-7pt}
\end{figure}

To implement this, a collection of tools is utilized. First, Google's Text to Speech (\texttt{gTTS})-API converts assigned text phrases to the robot. For instance, to call Alexa, the robot would utter a phrase such as "Alexa, what is the weather today?" to inquire about the weather. The synthesized speech generated by gTTS is played via a media player python library dubbed \texttt{playsound}. Alexa would naturally respond by stating the weather. For the robot to capture Alexa's voice utterance, it employs the Speech Recognition engine, which translates the voice expression to text through Google's Speech-to-Text API. This captured text phrase is forwarded to an Amazon Lex chatbot for interpretation. Amazon Lex is an easy-to-use conversational bot development tool, part of the AWS suite, capable of recognizing the intent of the text. According to this intent, the robot would utter back a statement to Alexa acknowledging its response and executes a certain action accordingly. Figure~\ref{weatherbot} illustrates the weather example we applied in the experiment if the weather is sunny. Upon acknowledgment of sunny weather, the robot navigates using the ROS \texttt{move\_base} package to a predefined set of coordinates; a "sunny" zone.

\section{Implementation and Experimental Setup} \label{sec:implementation}

\begin{figure}[H]
    \centering
    \includegraphics[width=.48\textwidth]{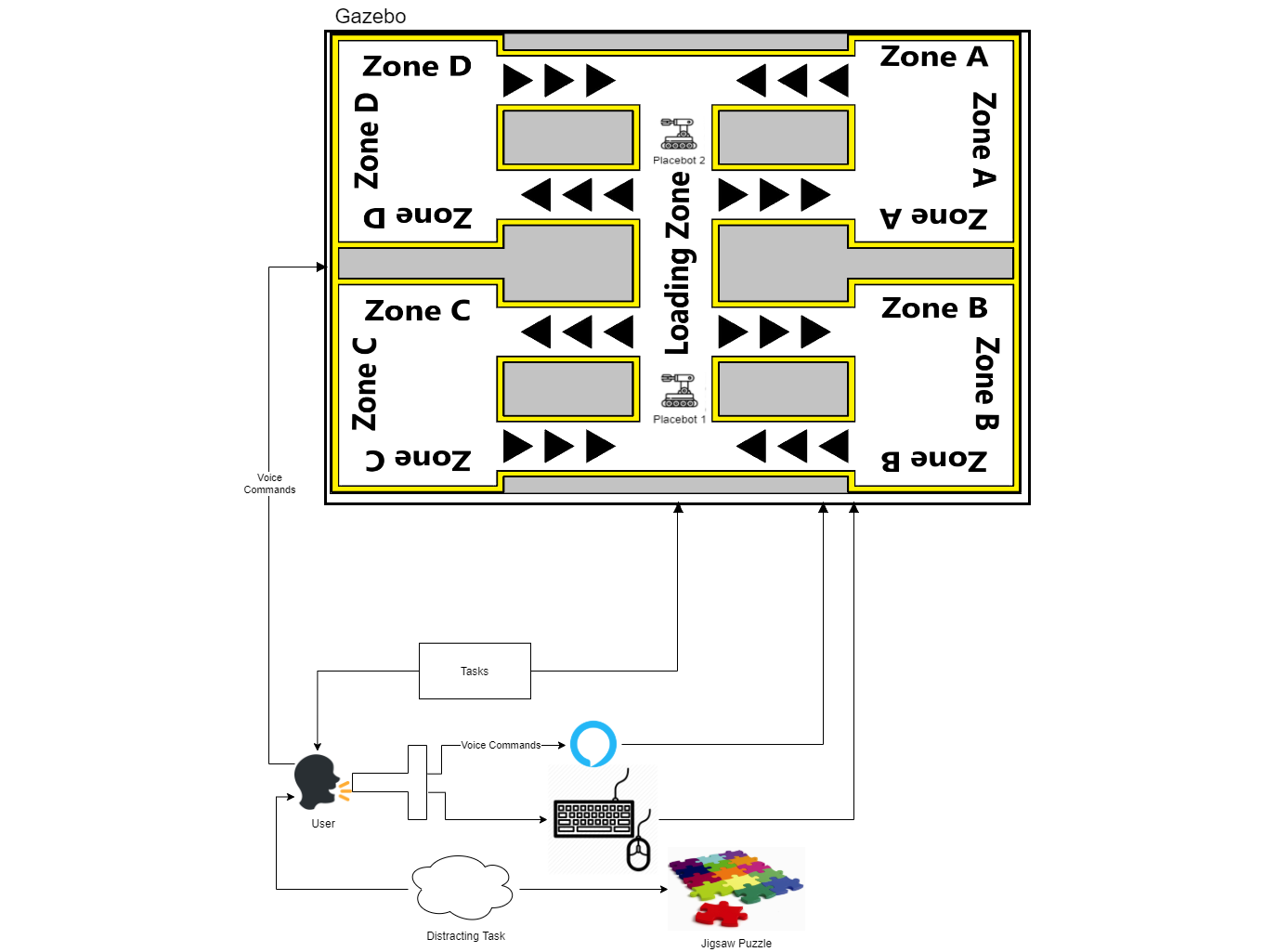}
    \caption{Experiment setup schematic}
    \label{virtual robot 10}
    \vspace{-14pt}
\end{figure}

We implemented our proposed design using Python and the AWS IoT SDK. The system was tested using an Amazon Alexa Echo dot and a mobile robot (placebot) running ROS melodic. A test setup was created to compare our approach to traditional robot control methods like a keyboard and mouse. A schematic diagram describing the testing environment is provided in Fig.~\ref{virtual robot 10}.

\begin{figure}[H]
    \centering
    \includegraphics[width=.47\textwidth]{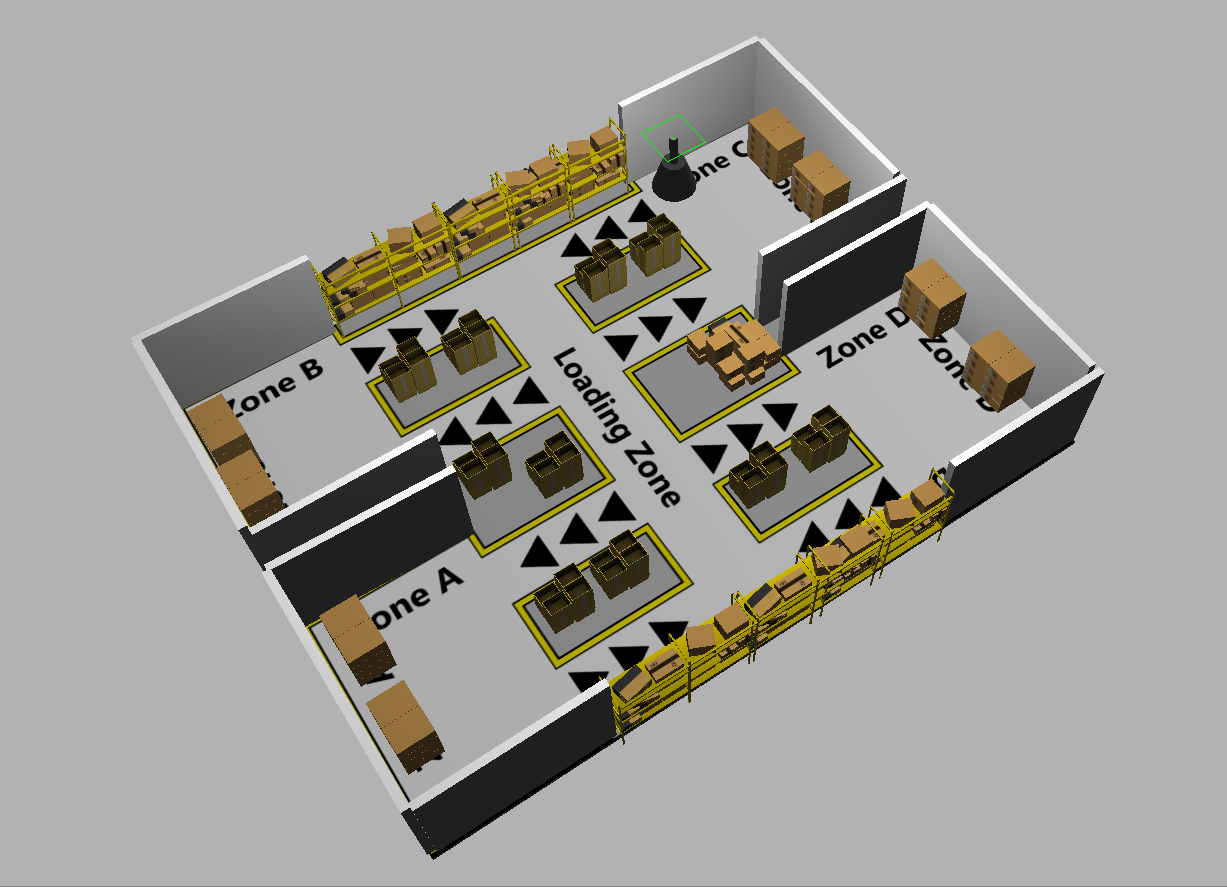}
    \caption{Gazebo world used for the experiment}
    \label{virtual robot 11}
    \vspace{-7pt}
\end{figure}

The experiment involves one test user controlling two robots and finishing several tasks along the way. The robots (placebots) are mobile robots programmed to be capable of autonomous navigation inside the allowed boundaries (white zones). The different robotic arms around the testing environment are also programmed to be able to pick and place color-coded (green) items from atop these mobile robots to the ground or from a conveyor belt (loading zone) to atop one placebot. These functionalities are essential, as the experiment's goal is to test our communication architecture and not robots' navigation or robot arm placement. Additionally, the testing environment is primarily virtual, with the robot environments simulated using ROS and Gazebo. This ensures an ideal and repeatable testing environment, especially for robot-to-Alexa communication, where a robot has to be near Alexa to ask and listen to responses. The robots are simulated on a computer on the same desk where Alexa is placed. The Gazebo world used is shown in Fig.~\ref{virtual robot 11}.

\begin{figure}[tbh]
    \centering
    \includegraphics[width=.45\textwidth]{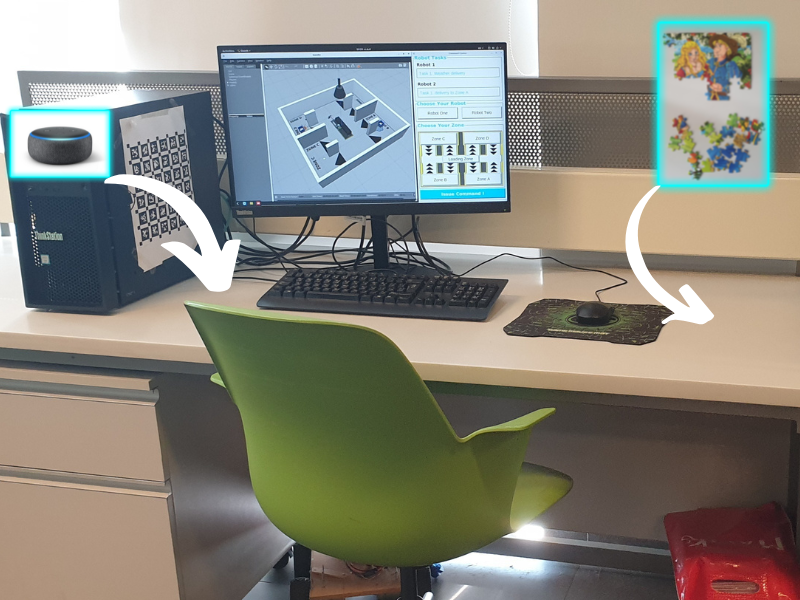}
    \caption{Experiment setup}
    \label{fig:experiment_setup}
    \vspace{-7pt}
\end{figure}

The user needs only to order the robots with different commands and solve a physical jigsaw puzzle. The complete setup is illustrated in Fig.~\ref{fig:experiment_setup}, showing the simulator and the location of the Amazon Echo Dot and the jigsaw puzzle, all within the user's reach on the bench. A total of 10 subjects (two females and eight males) participated in the experiment. Participants' ages ranged from 18 to 30 years. The flow of the experiment is explained below.
\begin{enumerate}
    \item The user will go through two runs for each control method, including one 5-minute warm-up run for the user to get used to the controls.
    \item A test run is deemed complete once all tasks are completed.
    \item The user is tasked with four tasks: three robot-related tasks and one distracting task.
    \item The distracting task entails the user finishing a jigsaw puzzle. No time limit is set on the tasks.
    \item The three robot tasks are designed based on functionality:
    
    \textbf{Task I - Alexa-to-Robot Communication:} In this task, a user orders the robots to move into the loading zone to be loaded with a package (green box), before ordering them to deliver the package to one of the Zones (A through D). The zone selection is randomized at the start of the experiment and is briefed to the user during the experiment (via a display on the computer). This task is deemed successful if each robot delivers its package to its set Zone. A failure is marked whenever a robot delivers a package to the wrong room. An intent is programmed in Alexa to enable the user to order a robot to navigate to one Zone (Zone A through D or Loading Zone). Robotic arms will load or unload the robots once they reach the zone waypoint.
        
    \textbf{Task II - Robot-to-Alexa Communication:} In this task, a user orders the robots to deliver a package based on the weather. The weather is pre-programmed to be randomized by the Alexa skill and is specific to the experiment (i.e., not the real weather). As such, the robots are expected to deliver packages to Zone A for sunny weather and Zone C for rainy weather. There are two ways of approaching this problem:
    
    \begin{itemize}
            \item \textbf{Traditional Control (keyboard and mouse):} the user has to ask Alexa for the weather twice (once for each robot) before ordering navigation to the correct zones. 
            \item \textbf{Proposed Communication Architecture:} the user can trigger a weather-navigation intent on Alexa. This will communicate to one robot that the user wants it to navigate based on the weather. Upon receiving the command, the robot will ask Alexa for the weather, interpret the response, and autonomously load, navigate, and unload a package to the correct zone. This will also test the robot's ability to communicate with Alexa through voice commands.
        \end{itemize}
    
    This task is deemed completely successful if each robot delivers one package to the correct zone (based on the weather). Failure in this task is defined as a robot navigating to the wrong room for the weather (i.e., Zone C when the weather is sunny). This is marked as a robot's failure to interpret Alexa's response correctly. An intent is programmed in Alexa to enable the user to order a robot to navigate based on the weather. Robotic arms will load or unload the robots once they reach the respective zone way-point.
        
    \textbf{Task III - Robot-to-Robot Communication:} In this task, the user is ordered to make the robots deliver two packages to Zone D in sequential order: placebot 1 delivers a package to Zone D, then placebot 2 delivers another package to zone D.  This task is deemed successful if each robot delivers one package to Zone D and placebot 1 delivers the package before placebot 2. If the task is successful, a green LED is turned on inside the simulation environment. The task is deemed failed if either of the robots fails to deliver the package to Zone D (i.e., delivered to a different Zone) or the order of delivering the packages is wrong (i.e., placebot 2 delivers the package before placebot 1). An intent is programmed in Alexa to enable the user to order a robot to start communicating with the other robot to coordinate sequential delivery. Robotic arms will load or unload the robots once they reach the respective zone way-point.
    \item Once the experiment is over (all four runs are finished), the user is given a NASA task load index (TLX) appended with two additional questions specific to our experiment.
\end{enumerate}

The evaluation criteria used during the experiment are split between subjective and objective metrics. The subjective metrics are filled in by the user after they are done with the experiment. The objective metrics are automatically collected and measured during the experiment. An observer only fills in the calculated results after each run. The metrics used during the experiment are summarized in Table~\ref{table_metrics}.

\begin{table}[h]
\caption{Summary of Evaluation Metrics}
\vspace{-12pt}
\label{table_metrics}
\begin{center}
\begin{tabular}{|c||l|}
\hline
\textbf{Metric Type} & \textbf{Parameter}\\
\hline
\multirow{8}{*}{Subjective} & Mental Demand\\

 & Physical Demand\\

 & Temporal Demand\\

 & Performance\\

 & Effort\\

 & Frustration\\

 & Confidence\\

 & Trust\\
\hline
\multirow{9}{*}{Objective} & Command Success Rate\\

 & Task Success Rate\\

 & Communication Success Rate\\

 & Command Failure Rate\\

 & Task Failure Rate\\

 & Communication Failure Rate\\

 & Robot Attention Demand (RAD)\\

 & Fan-out (FO)\\

 & Time Taken\\
\hline
\end{tabular}
\vspace{-14pt}
\end{center}
\end{table}

\section{Results and Analysis}
\label{sec:results}
Objective measures were used to evaluate the ability to communicate with and command the robots while solving the puzzle. Both methods achieved 100\% command, task, and communication success rates across all trials and for all test subjects. Four objective metrics (RAD, FO, time taken, and the number of solved puzzle pieces) were used to quantitatively compare the effect of both control methods on the users’ performance while completing the assigned tasks. 

The results are summarized in Table~\ref{table:objective_results} per method showing the average score across both trials. The VAHR method required nearly 41\% less Robot Attention Demand while ensuring almost 93\% more Fan-out time. This shows significant improvement in controlling both robots simultaneously without the need for increased attention from the user. Although users needed more time to complete the task using VAHR (254 seconds) compared to the standard mouse-and-keyboard method (234 seconds), VAHR allowed the users to solve more pieces of the puzzle averaging a 62.5\% improvement. An ANOVA test was conducted to assess the statistical significance of the obtained objective results. The p-values obtained from comparing all metrics under both control methods are listed in Table~~\ref{table:objective_results}. The ANOVA test results show strong evidence against the null hypothesis (i.e., $p-values < \alpha=0.05$), thus proving that the results are statistically significant.

\begin{table}[tbhp]
    \vspace{-7pt}
    \caption{Objective Results}
    \vspace{-7pt}
    \centering
    \begin{tabular}{lcccc} \toprule
        & VAHR & Mouse \& Keyboard & p-value \\ \midrule
    Interaction Time (s) & 111 & 175 &\textbf{ 2.9e-4} \\
    Neglect Time (s) & 142 & 59 &\textbf{ 2e-5}\\
    RAD [0, 1] & 0.44 & 0.75 & \textbf{7.3e-7}\\ 
    Fan-out (s) & 578 & 314 & \textbf{6.3e-4}\\ 
    Solved Puzzle Parts & 13 & 8 & \textbf{3e-2}\\ \bottomrule
    \end{tabular}
        \label{table:objective_results}
    \vspace{-7pt}
\end{table}

\begin{figure}[tbhp]
    \centering
    \includegraphics[width=0.45\textwidth]{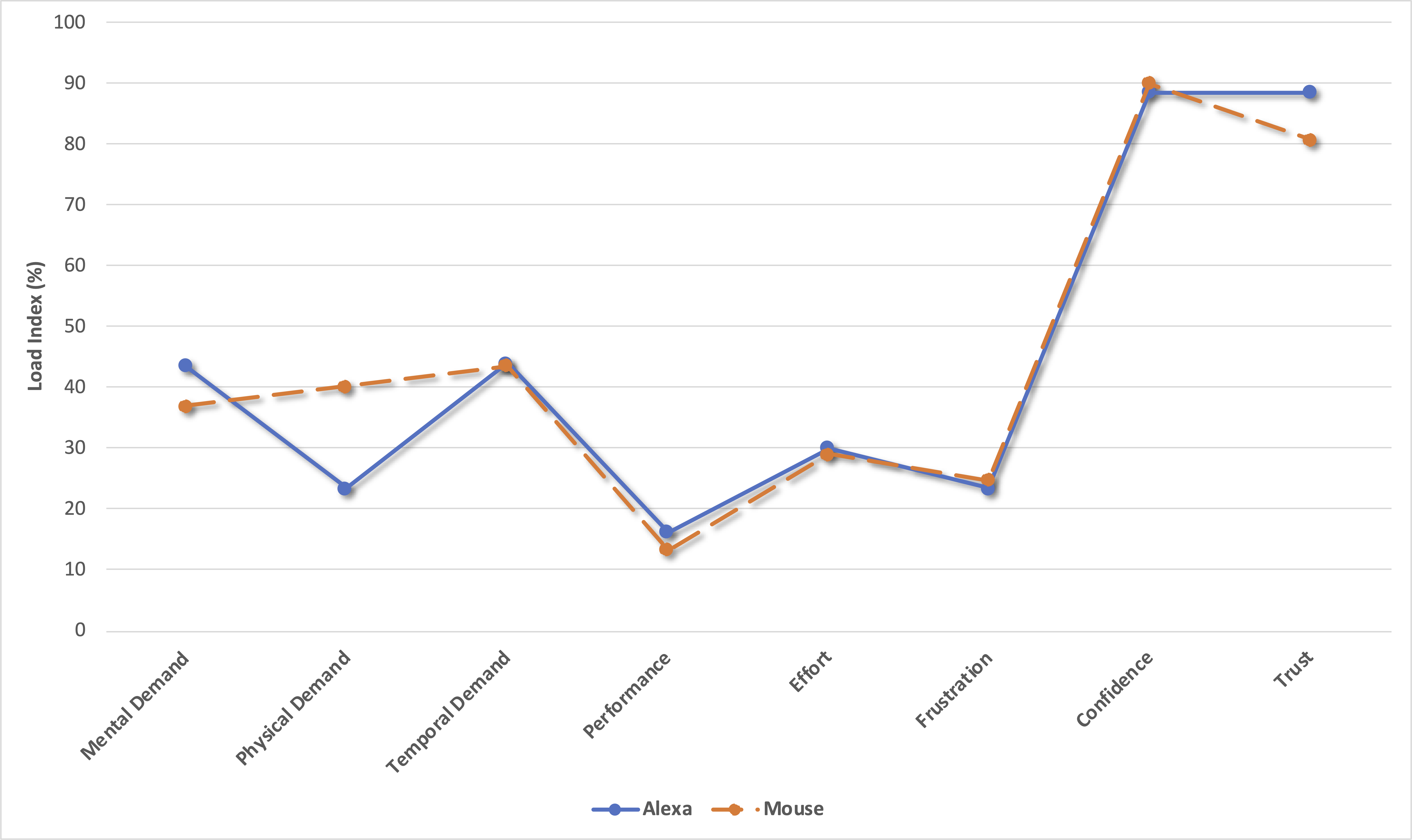}
    \caption{Subjective results comparison}
    \label{fig:subjective_results}
\end{figure}

Figure~\ref{fig:subjective_results} shows the averages of the NASA Task Load Index subscales. The NASA Task Load Index consists of a scale of 21 assessment levels which we converted to a percentile score from 0\% at level 1 to 100\% at level 21. The lower score reflects less load on the subject under testing. While VAHR required 17.6\% more mental demand on average than the traditional communication method, the physical load on the subject was decreased by 41.6\%. The temporal load was almost the same for both methods with a 1\% difference. The users assess their own performance to be 22.6\% better when using the traditional method which is also reflected in the 9.6\% more trust. The subjective assessment of performance contradicts the objective evaluation scores as the subjects show superior performance while completing the tasks using VAHR. We estimate that the discrepancy goes back to the fact that the proposed VAHR method is relatively new to the subjects who need time to build trust in the new control method. VAHR required 3.5\% more effort on average from the users to complete the assigned tasks but decreased the frustration level by 5.1\% and improved the subjects’ confidence in the ability to complete the tasks successfully by 1.7\%. The ANOVA test conducted on the subjective results obtained from both control methods resulted in a p-value of 0.28 ($p-value<\alpha=0.05$) which doesn't show sufficient statistical significance for the obtained results. Individual ANOVA tests were conducted on the results of each subjective metric of the NASA TLX to check for statistical significance. None of the subjective results received a p-value lower than 0.05 across the two methods and therefore the null hypothesis cannot be rejected for any of the subjective metrics.

\begin{figure}[tbhp]
    \centering
    \includegraphics[width=0.45\textwidth]{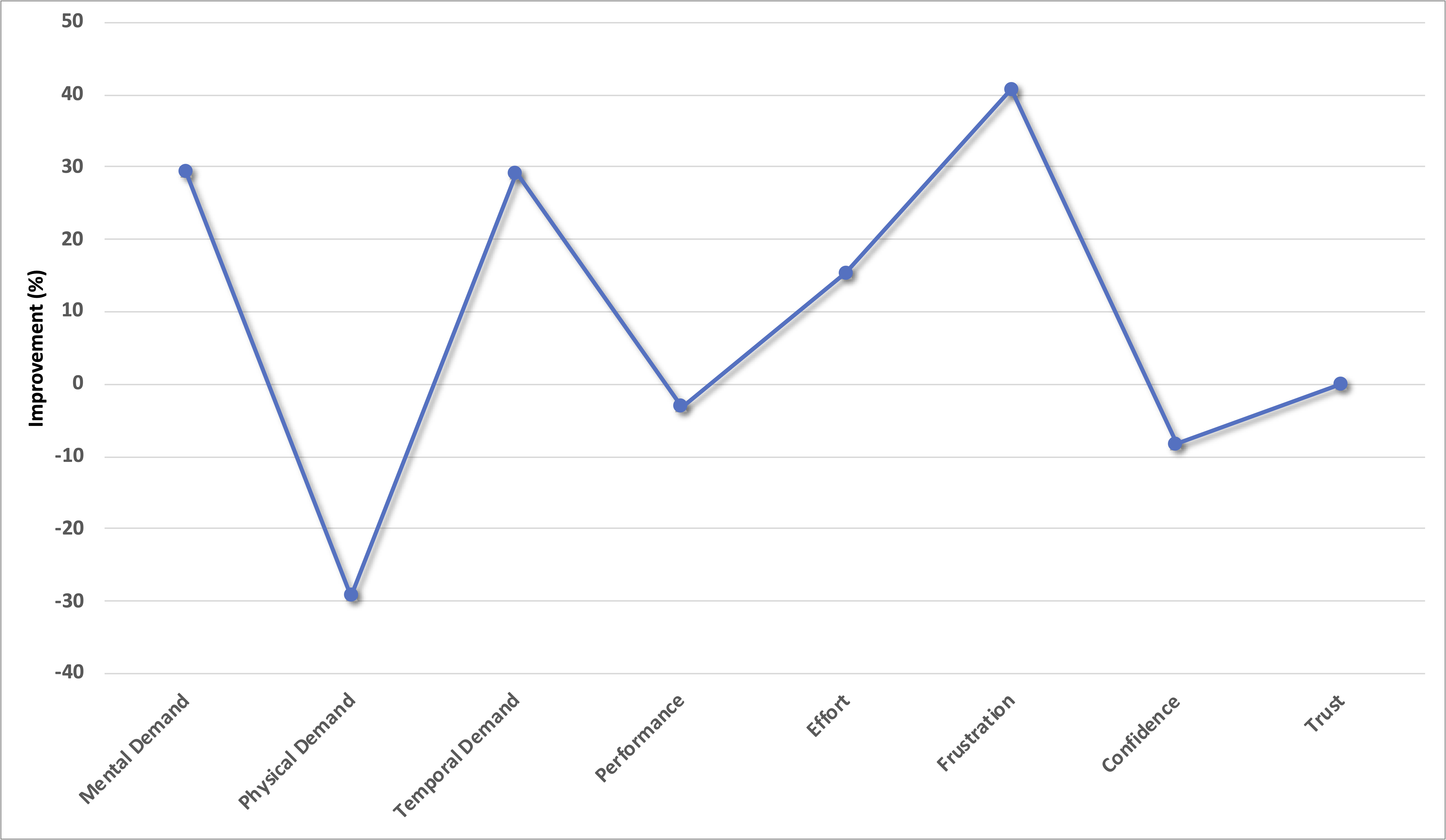}
    \caption{VAHR - subjective results improvement across trials}
    \label{fig:subjective_results_alexa_imprv}
    \vspace{-14pt}
\end{figure}

Since VAHR is relatively new to the users, we consider the change in the NASA Task Load Index scores across different trials. As shown in Fig.~\ref{fig:subjective_results_alexa_imprv}, the users felt that the task required less mental and temporal demand in the second trial compared to the first one, with each of the two metrics registering an improvement of around 29\%. Improvement is also evident in the effort index, with around 15\% and the highest percentage improvement is in the frustration index which reaches more than 40\% in the second trial. On the other hand, VAHR required 29\% more physical demand on average from the users in their second trial where they also felt that their performance and confidence dropped by around 3\% and 8\% respectively. The trust score remained constant across both trials.

\section{Conclusion}
\label{sec:conclusion}
We present VAHR, a scalable cloud-based communication paradigm that connects humans to robots through virtual assistants. This communication system enables all agents to communicate freely with each other through speech. While it requires more setup than traditional approaches, cloud services provide increased robustness and scalability. We test our system against standard mouse-and-keyboard control through a set of human-subject experiments, which demonstrated the advantage of using VAHR in objective metrics, providing at least 41\% improvement. Part of the subjective measures noted by users shows a preference for mouse-and-keyboard control, which we attribute to unfamiliarity with this novel communication scheme. However, the high p-values obtained from the ANOVA tests on both combined and individual subjective metrics show low statistical significance of the subjective test results. This adds evidence to the unfamiliarity of the subjects with the proposed control/communication method, even from a self-assessment point of view. We suggest that our system would be best suited for environments such as warehouses and hospitals, where physical tasks have to be balanced with robot control.

\bibliographystyle{IEEEtran}
\bibliography{references}{}

\end{document}